\documentclass[letterpaper, 10 pt, conference]{ieeeconf}  

\IEEEoverridecommandlockouts                              

\usepackage{graphics} 
\usepackage{graphicx}
\usepackage{epsfig} 
\usepackage{mathptmx} 
\usepackage{times} 
\usepackage{amsmath} 
\usepackage{amssymb}  
\usepackage{booktabs}
\usepackage{makecell}
\usepackage{xcolor} 
\usepackage{multirow}
\usepackage{amssymb}
\usepackage{ragged2e}
\usepackage{pifont}
\usepackage{subcaption}
\newcommand\our{Uni-Skill}
\newcommand\ourdata{SkillFolder}

\title{\LARGE \bf
\our{}: Building Self-Evolving Skill Repository for Generalizable Robotic Manipulation
}

\author{Senwei Xie, Yuntian Zhang, Ruiping Wang and Xilin Chen
\thanks{*This work is partially supported by Beijing Municipal Natural Science Foundation Nos. L257009, L242025, and Natural Science Foundation of China under contracts Nos. 62495082, 62461160331.}
\thanks{The authors are with Key Laboratory of AI Safety of CAS, Institute of Computing Technology, Chinese Academy of Sciences, Beijing, 100190, China, and with University of Chinese Academy of Sciences, Beijing, 100049, China. \tt\footnotesize\justifying{\{senwei.xie,yuntian.zhang\}} \tt\footnotesize\justifying{@vipl.ict.ac.cn, \{wangruiping,xlchen\}@ict.ac.cn}.}
\thanks{Corresponding Author: Ruiping Wang.}
}

\begin{document}

\maketitle
\thispagestyle{empty}
\pagestyle{empty}

\begin{abstract}

While skill-centric approaches leverage foundation models to enhance generalization in compositional tasks, they often rely on fixed skill libraries, limiting adaptability to new tasks without manual intervention. To address this, we propose \our{}, a \textbf{Uni}fied \textbf{Skill}-centric framework that supports skill-aware planning and facilitates automatic skill evolution. Unlike prior methods that restrict planning to predefined skills, \our{} requests for new skill implementations when existing ones are insufficient, ensuring adaptable planning with self-augmented skill library. To support automatic implementation of diverse skills requested by the planning module, we construct \ourdata{}, a VerbNet-inspired repository derived from large-scale unstructured robotic videos. \ourdata{} introduces a hierarchical skill taxonomy that captures diverse skill descriptions at multiple levels of abstraction. By populating this taxonomy with large-scale, automatically annotated demonstrations, \our{} shifts the paradigm of skill acquisition from inefficient manual annotation to efficient offline structural retrieval. Retrieved examples provide semantic supervision over behavior patterns and fine-grained references for spatial trajectories, enabling few-shot skill inference without deployment-time demonstrations. Comprehensive experiments in both simulation and real-world settings verify the state-of-the-art performance of \our{} over existing VLM-based skill-centric approaches, highlighting its advanced reasoning capabilities and strong zero-shot generalization across a wide range of novel tasks.

\end{abstract}

\begin{figure*}[t]
      \centering
      \includegraphics[width=0.95\textwidth]{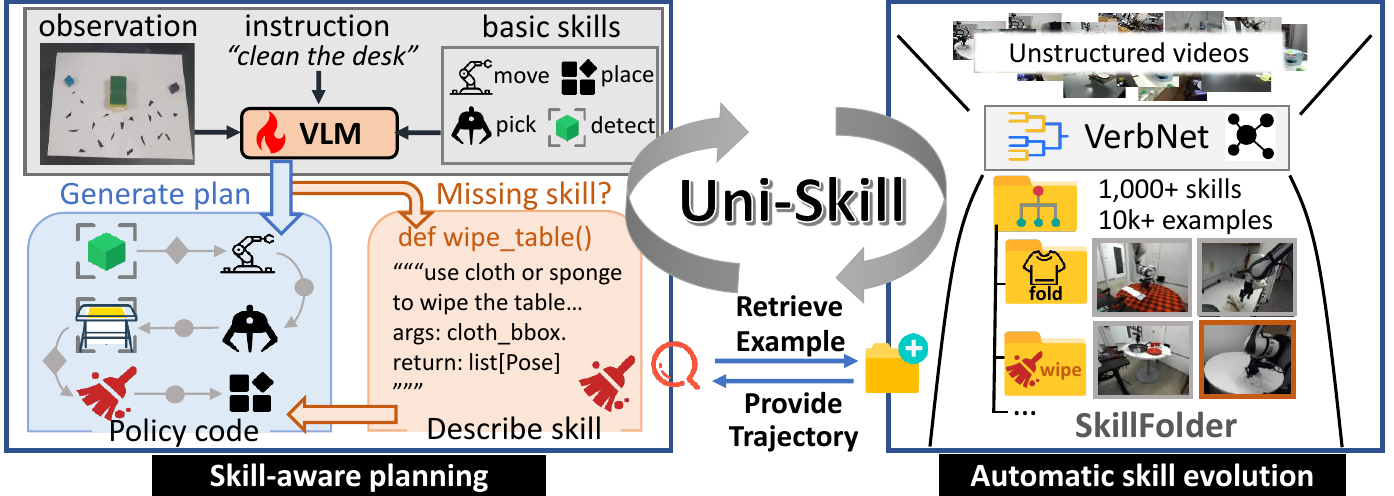}
    \caption{\our{} facilitates the generalization of the planning process beyond predefined skills with self-augmented skill descriptions. By organizing extensive unstructured robotic videos with the hierarchical skill repository, \ourdata{}, we enable efficient retrieval of skill demonstrations and automatic implementation of newly defined skills at deployment.} 
    \vspace{-1em}
    \label{fig:overall}
   \end{figure*}
\section{INTRODUCTION}

Following free-form language instructions to accomplish diverse real-world tasks is a long-term goal in robotic learning. This challenge requires robots to adapt to novel instructions while generalizing across different environments and contexts. End-to-end behavior cloning methods~\cite{rt2, openvla, llarva, openvla-oft} benefit from large-scale demonstrations and show high precision on in-distribution tasks. However, this task-oriented approach requires additional fine-tuning for new environments and tasks. Different from task-oriented methods using end-to-end action outputs, skill-centric approaches~\cite{cap,instruct2act, robocodex} formulate manipulation tasks more hierarchically. Complex language instructions are decomposed into pre-defined skills, leveraging the generalization ability of pre-trained large language models (LLMs) as code planners. 

However, existing skill-centric approaches are inherently constrained by a fixed skill set. If an API for \textit{fold clothes} is unavailable, the system is fundamentally incapable of executing this task variation. Even when augmented with textual skill descriptions, existing methods still rely heavily on manually curated demonstrations or waypoint annotations at deployment, requiring additional supervision for each novel skill~\cite{parakh2024lifelong}. While large scale robotic videos inherently contain promising source as skill demonstrations, these unstructured and noisy data lack the necessary links and annotations to specific skills, making them unsuitable for direct use in the expansion and supplementation of skills. 

These limitations expose a deeper issue: current frameworks lack the ability to identify and adapt to skill gaps, and fail to effectively establish intrinsic connections between diverse skills and potential data sources. Addressing this calls for a shift in perspective: skills should no longer be viewed as fixed, isolated API primitives, but as expandable, evolving entities that can form structured associations with broader data sources, enabling automated acquisition at deployment. This motivates two core capabilities: (1) \textbf{skill-aware planning}, where the system detects missing skills and generates high-level descriptions to support adaptive decomposition; and (2) \textbf{automatic skill evolution}, which reduces human supervision by efficiently linking large-scale, unstructured robotic data to diverse skill implementations. We integrate both components into \textbf{\our{}}, a \textbf{Uni}fied \textbf{Skill}-centric framework to support more scalable skill-centric robot learning across diverse real-world scenarios.

The \textbf{skill-aware planning} module is designed to generalize the planning process beyond a predefined skill library, enabling it to handle novel task decompositions. As illustrated in Fig.~\ref{fig:overall}, given a set of basic skills, \our{} first evaluates whether these fundamental skills are sufficient to execute the given instruction like \textit{clean the desk}. If additional capabilities are required, \our{} autonomously generates descriptions for supplementary skills. On the one hand, these additional skill descriptions formally extend the original skill repository, ensuring planning adaptability to tasks beyond the scope of pre-defined skill sets. On the other hand, they serve as semantic anchors to match with unstructured demonstrations during skill implementations.


The \textbf{automatic skill evolution} module is designed to ground high-level skill descriptions, as requested by the planning module, into low-level, skill-centric action sequences. Rather than relying on manually collected demonstrations at deployment, we explore the potential of large-scale, unstructured robotic videos. In analogy to how ImageNet~\cite{deng2009imagenet} leveraged WordNet~\cite{miller1995wordnet} to structure visual object categories, we devise \ourdata{}, a VerbNet-inspired dataset capturing hierarchical skill categories~\cite{schuler2005verbnet}. Building upon this skill-centric hierarchy, video segments with automatically annotated descriptions are iteratively incorporated to expand and enrich each layer of \ourdata{}, yielding a densely annotated dataset of over 10,000 skill traces, which are mapped to 106 VerbNet classes and 1,659 unique skill formulations. By doing so, demonstrations for newly defined skills can be efficiently retrieved from semantically relevant examples in \ourdata{}. These retrieved skill traces further support few-shot inference by providing skill-centric behavior patterns and spatial trajectory references, facilitating automatic implementation without deployment-time demonstrations.

Extensive experiments on simulation and real-world environments confirm state-of-the-art zero-shot performance of \our{} on diverse manipulation tasks. For tasks out of the predefined skills on RLBench~\cite{rlbench}, the zero-shot success rate of \our{} outperforms the state-of-the-art visual prompting method MOKA~\cite{liu2024moka} by 31.0\%. On real-world settings, the performance of \our{} achieves an improvement of 20.0\% on long-horizon tasks and 34.0\% in unseen skills. 

\begin{figure*}[t]
    \centering
    \includegraphics[width=0.95\textwidth]{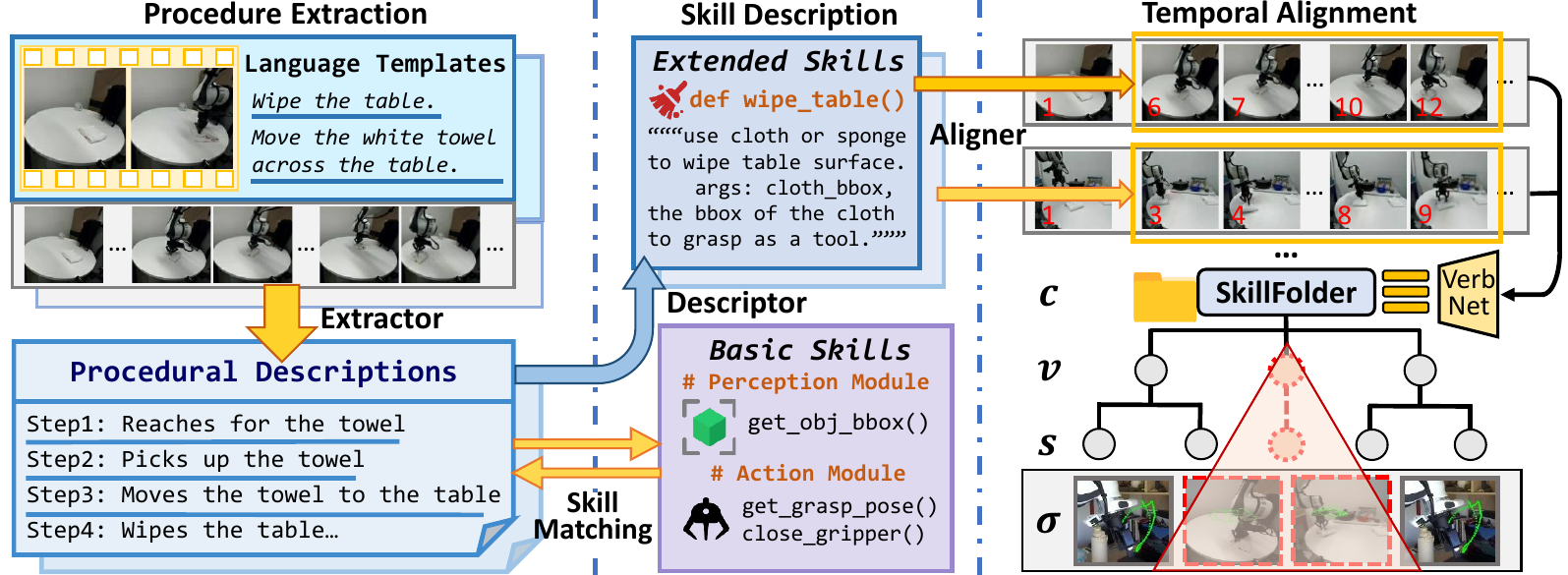}
    \caption{The \textbf{automatic skill annotation} pipeline. Skill functions derived from procedural descriptions are aligned with the video segments and iteratively involved in the construction of \ourdata{}.}
    \vspace{-1em}
    \label{fig:data}
\end{figure*}

\section{RELATED WORKS}
\textbf{Robotic task decomposition.}
Generalizing robotic models to long-horizon complex tasks is a longstanding challenge in robotic manipulation. A common strategy is to decompose long-horizon tasks into subtasks and atomic behaviors, reducing execution complexity and enhancing compositional generalization~\cite{ogren2022behavior, marzinotto2014towards}. Recent advances in large language models (LLMs) have significantly accelerated progress in robotic task planning. Leveraging LLMs, free-form language instructions can be decomposed into natural language plans in a zero-shot manner~\cite{saycan, palme, inner}. To bridge the gap between high-level plans and low-level execution, policy code generation approaches abstract atomic skills into code APIs, enabling the generation of executable code that parameterizes predefined skills~\cite{cap,instruct2act,robocodex,xie2025robotic,progprompt,pmlr-v162-huang22a,huang2023voxposer}. Compared with end-to-end methods, these skill-centric approaches emphasize the reusability of atomic skills across diverse tasks and exploit their inherent compositionality.

\textbf{Construction of skill repository.}
For skill-centric methods, the capacity of the skill library is a basis for generalizing to diverse tasks and environments. Policy code generation methods primarily construct skill libraries with a fixed set of parameterized APIs, which restricts the ability to generalize beyond the predefined skills~\cite{cap,instruct2act,progprompt,creative,vemprala2024chatgpt}. Some LLM-based approaches leverage human feedback to determine whether new skills are required~\cite{parakh2024lifelong}. New skills can be taught through interactive spoken dialogue and kinesthetic demonstrations~\cite{grannen2025vocal,grannen2025provox}. However, skill acquisition still relies heavily on human feedback and manual demonstrations during deployment. In contrast, a large amount of robotic videos already exist, covering diverse skills but lacking explicit annotations. If these unstructured videos can be organized with hierarchical skill labels, skill-relevant examples could be efficiently retrieved, enabling more scalable and cost-effective expansion of the skill library.

\textbf{Behavior retrieval.}
A substantial body of research has investigated retrieval-based approaches for robotic manipulation~\cite{malato2024zero, di2024dinobot}. Retrieval-based behavior cloning methods match large-scale unstructured demonstrations with few-shot task-specific data, scaling policy training with retrieved examples~\cite{nasiriany2023learning, du2023behavior, memmelstrap}. However, these methods typically focus on instance-level matching rather than organizing demonstrations by reusable skills. As a result, demonstrations remain loosely structured and non-hierarchical, leading to repeated feature-matching during retrieval and limited compositional generalization. In contrast, Uni-Skill adopts a skill-centric perspective, hierarchically structuring demonstrations through \ourdata{}, and enabling direct retrieval via semantic labels without additional deployment-time curation.

\section{METHOD}

\subsection{Skill-Aware Planning}
\label{sec:plan}
We consider language-conditioned robotic manipulation tasks with multimodal inputs, where each task is instructed with a free-form language instruction $I_t$ and corresponding visual observations $O_t$. Skill-centric methods are built on the assumption of a pre-defined skill library $L_{API}$. Fundamental atomic skills, such as pick and place, are more frequently discussed and typically implemented in the basic skill library $L_{\text{base}}$. In scenarios that require more intricate motion planning, such as \textit{fold cloth}, compositions of only basic skills often prove insufficient. Rather than statically encoding all possible task-specific behaviors, a more effective strategy is to allow the repository to dynamically expand in response to the instruction $I_t$. This results in a hybrid skill set $\{L_{\text{base}}, L_{\text{ext}}\}$, where $L_{\text{ext}}$ is synthesized or retrieved on demand, ensuring sufficiency for specific task requirements.

The process of adaptively extending skill descriptions to generate task-compliant plans is referred to as \textbf{skill-aware planning}. This paradigm requires the system to evaluate whether existing skills sufficiently fulfill the task requirements, synthesize additional skill descriptions, and generate executable plans conditioned with self-augmented skills. We formalize these interdependent capabilities into three core modules: the sufficiency discriminator $\mathcal{E}$, the skill generator $\mathcal{G}$, and the planner $\mathcal{P}$. As shown in Fig.~\ref{fig:overall}, given the API descriptions of a basic skill set $L_\text{base}$ and the multimodal instruction pair $\{O_t, I_t\}$, the discriminator $\mathcal{E}$ first assesses whether the available basic skills are sufficient to execute the given instruction, providing indicative feedback denoted as $\mathcal{E}(\{O_t,I_t\}, L_\text{base})$. If additional capabilities required, the skill generator $\mathcal{G}$ autonomously synthesizes new skills $\mathcal{G}(\{O_t,I_t\}, L_{\text{base}})$ to supplement existing ones. 

With the visual observation and language instruction as inputs, the planner $\mathcal{P}$ generates executable policy code $\{\pi_i, p_i\}_{i=1}^N$ conditioned on the self-augmented API library, where each $\pi_i$ denotes the $i$-th API call from either the base set $L_{\text{base}}$ or the extended set $L_{\text{ext}}$, and $p_i$ represents the corresponding parameters. The entire generation process can be formulated as:
\begin{equation}
	(O_t, I_t, \{L_\text{base}, L_\text{ext}\}) \overset{\mathcal{P}}{\Longrightarrow} \{\pi_i, p_i\}_{i=1}^N.
\end{equation}
Notably, the three key components, $\mathcal{E}$, $\mathcal{G}$, and $\mathcal{P}$ share a unified, code-based output format and support multi-modal inputs, aligning well with the pipeline of large vision-language models (VLMs). With their comprehensive contextual understanding and advanced visual grounding capabilities, VLMs can serve as intelligent planners and creative skill descriptors. To adapt the VLM to robotic manipulation tasks in a code-based framework, we trained it with 106K runtime code samples derived from videos demonstrations. 

\subsection{Automatic Skill Evolution}
To enable automatic grounding of extended skill descriptions into executable actions, we devise the \textbf{automatic skill evolution} module. This framework systematically connects diverse skill semantics to unstructured demonstrations, forming a reusable repository for few-shot skill implementation without manual intervention. It consists of three structured components implemented in a progressive order: \textbf{automatic skill annotation} for unstructured videos, \textbf{hierarchical skill organization} using VerbNet, and \textbf{few-shot skill implementation} with efficient example retrieval.

\begin{figure*}[t]
    \centering
    \includegraphics[width=0.98\textwidth]{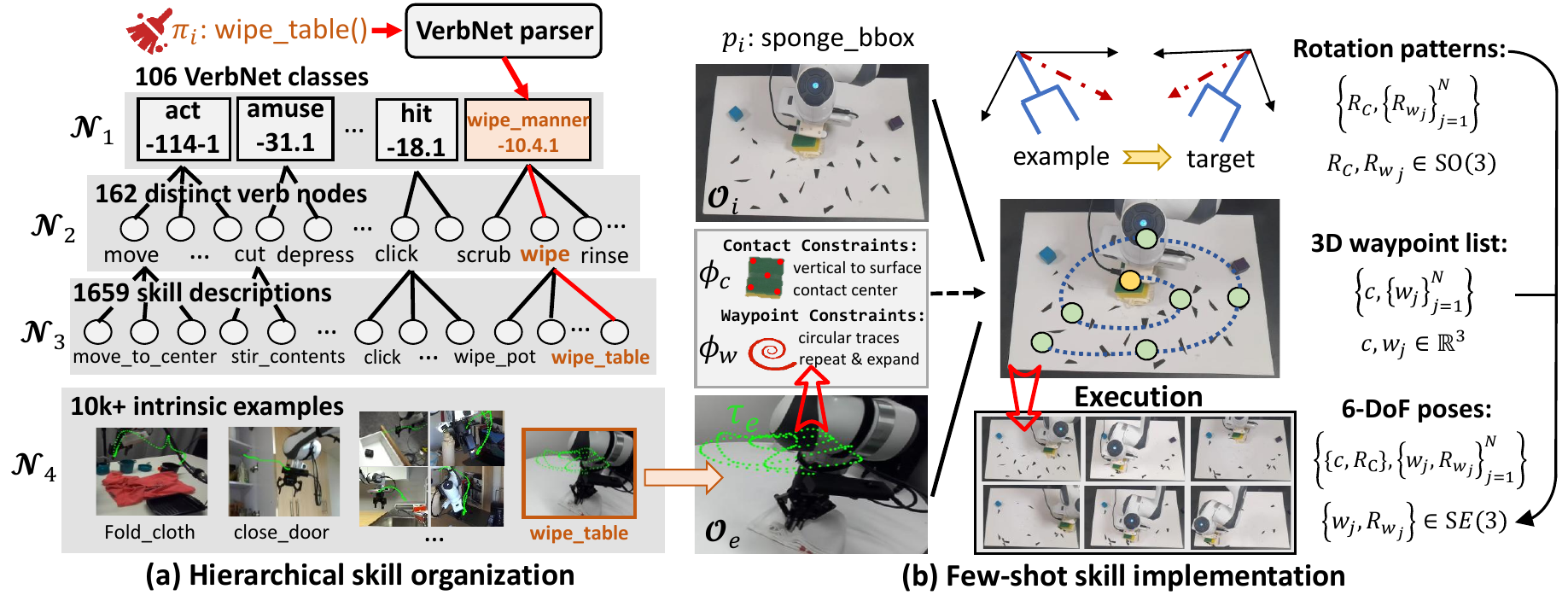}
    \caption{(a) Skill examples are retrieved from \ourdata{} hierarchically at deployment. (b) Semantic constraints and action flows serve as in-context examples for waypoints generation, which are further lifted to 6-DoF poses with rotational patterns.}
    \vspace{-1.0em}
    \label{fig:skill}
\end{figure*}

\paragraph{Automatic skill annotation.}
To address high cost and low efficiency of collecting new demonstrations at deployment, we draw inspiration from abundant demonstration videos in-the-wild. These robotic videos inherently encode procedural knowledge across diverse skill implementations. However, raw videos are often unstructured, lacking explicit skill annotations and containing irrelevant segments, which limits their utility as direct, skill-centric demonstrations. To convert these potential resources into skill-centric formats, we propose a VLM-based pipeline that automatically segments videos and labels each clip with the corresponding skill descriptions. As shown in Fig.~\ref{fig:data}, this pipeline consists of three stages: procedure extraction, skill description, and temporal alignment. Accordingly, we prompt Gemini-2.0-Flash~\cite{gemini} to perform three dedicated roles: the Extractor for procedure extraction, the Descriptor for skill description, and the Aligner for temporal alignment.

In the first stage, the Extractor distills procedural knowledge from raw videos by generating concise action plans. These plans are conditioned on keyframes at physical transition points~\cite{wang2024vlmseerobotdo}. Outputs from multiple instruction templates are consolidated into unified, step-by-step descriptions, which serve as structured priors for subsequent stages. In the second stage, based on these descriptions, we extract task-oriented skill descriptions aligned with the demonstrations. To ensure consistency and avoid redundant definitions, the Descriptor is given a basic skill set and checks whether it suffices for the step-by-step procedure. If necessary, the Descriptor further synthesizes new task-oriented skill descriptions, adhering to conventions of the basic skills. In the third stage, we further align these task-oriented skill annotations with specific intervals of the video demonstrations. To better capture the temporal relationships in videos, we employ the Number-It~\cite{numberit} strategy, explicitly annotating each keyframe with a fixed-size integer to indicate its sequential order. The Aligner is then prompted to select the integer interval that best corresponds to the target skill description. In this manner, we extract over 10,000 skill-centric video segments from 350 hours demonstrations of DROID~\cite{droid}, converting unstructured robotic videos to skill-annotated example slices.


\begin{table*}[h]
    \caption{Success rate on 8 RLBench tasks covered by the basic skills.}
        \vspace{-1.0em}
    \setlength{\tabcolsep}{5.0pt}
    \label{tab:rlbench}
    \begin{center}
    \begin{tabular}{l|c|ccccccccccc}
    \toprule
    \bf Models &  setting &\makecell{Push\\Buttons} &  \makecell{Stack\\Blocks} & \makecell{Close\\Jar}  &  \makecell{Stack\\Cups} &  \makecell{Sweep\\Dirt} &  \makecell{Slide\\Block} &  \makecell{Screw\\Bulb} &  \makecell{Put in\\Board} &  Avg. \\
    \midrule 
    CaP (GPT-3.5)~\cite{cap}& ICL & 0.76$\pm0.04$ & 0.12$\pm0.11$ & 0.00$\pm0.00$ & 0.00$\pm0.00$ & 0.43$\pm0.14$ & 0.04$\pm0.00$ & 0.03$\pm0.02$ & 0.04$\pm0.04$ & 0.18$\pm0.01$\\
    CaP (GPT-4o)~\cite{gpt4o}& ICL & \bf 0.84$\pm0.00$ & 0.31$\pm0.06$ & 0.03$\pm0.02$ & 0.01$\pm0.02$ & \bf 0.64$\pm0.00$ & 0.21$\pm0.02$ & 0.13$\pm0.02$ & \bf 0.08$\pm0.04$ & 0.28$\pm0.02$\\
    \bf \our{} (ours)& ZSL & 0.77$\pm0.02$ & \bf 0.39$\pm0.05$ & \bf 0.48$\pm0.04$ & \bf 0.05$\pm0.02$ & \bf 0.64$\pm0.00$ & \bf 0.67$\pm0.06$ & \bf 0.31$\pm0.02$ & \bf 0.08$\pm0.04$ & \bf 0.42$\pm0.01$\\
    \bottomrule
    \end{tabular}
    \end{center}
\end{table*}
\begin{table*}[h]
    \caption{Success rate on 10 RLBench tasks out of the basic skill set.}
    \vspace{-1.5em}
    \setlength{\tabcolsep}{3.0pt}
    \label{tab:rlbench_extra}
    \begin{center}
    \begin{tabular}{l|cccccccccccc}
    \toprule
    \bf Models &  \makecell{Close\\Micro} &  \makecell{Close\\Fridge} &  \makecell{Seat\\Down} &  \makecell{Close\\Laptop}  &  \makecell{Close\\Drawer} &  \makecell{Press\\Switch} &  \makecell{Water\\Plants} &  \makecell{Open\\Door} &  \makecell{Unplug\\Charger} &  \makecell{Lift\\Number} &  Avg. \\
    \midrule
    CaP~\cite{cap} & 0.00$\pm0.00$ & 0.00$\pm0.00$ & 0.09$\pm0.05$ & 0.04$\pm0.04$ & 0.00$\pm0.00$ & 0.00$\pm0.00$ & 0.00$\pm0.00$ & 0.00$\pm0.00$ & 0.00$\pm0.00$ & 0.00$\pm0.00$ & 0.01$\pm$0.01\\
    MOKA~\cite{liu2024moka} & 0.05$\pm0.02$ & 0.17$\pm0.06$ & 0.23$\pm0.08$ & 0.07$\pm0.05$ & 0.19$\pm0.05$ & 0.33$\pm0.06$ & 0.00$\pm0.00$ & 0.00$\pm0.00$ & 0.00$\pm0.00$ & 0.00$\pm0.00$ & 0.10$\pm0.02$\\
    \bf \our{} (ours) & \textbf{0.49}$\pm0.06$ & \textbf{0.57}$\pm0.05$ & \textbf{0.68}$\pm0.04$ & \textbf{0.33}$\pm0.02$ & \textbf{0.56}$\pm0.00$ & \textbf{0.40}$\pm0.04$ & \textbf{0.05}$\pm0.02$ & \textbf{0.33}$\pm0.02$ & \textbf{0.39}$\pm0.02$ & \textbf{0.31}$\pm0.02$ & \textbf{0.41}$\pm0.01$\\
    \bottomrule
    \end{tabular}
    \vspace{-1.5em}
    \end{center}
\end{table*}

\paragraph{Hierarchical skill organization.}
Through automated skill annotation, we have obtained a large corpus of skill-annotated demonstrations. Despite skills are often treated as flat and isolated entities in existing methods, they inherently exhibit varying granularity and hierarchical relationships. Capturing this intrinsic structure enables more efficient sample retrieval and supports compositional generalization. Inspired by how ImageNet leverages WordNet to construct visual object categories, we adopt VerbNet as a foundation for structured skill categories. Unlike object categories, which are typically noun-based, skills encompass both actions and their thematic roles, necessitating finer-grained semantic modeling. Building on this foundation, we extend the abstract verb categories in VerbNet with additional layers that encode concrete skill realizations and exemplar visual scenes. This results in a four-layer skill tree, \ourdata{}, organizing diverse skill semantics into progressively refined skill taxonomy. The nodes \(\mathcal{N}\) of this hierarchy are partitioned into distinct levels, ranging from high-level verb classes to specific, visually grounded skill instances.

As shown in Fig.~\ref{fig:skill} (a), the top layer \(\mathcal{N}_1\) corresponds to distinct VerbNet classes, such as \textit{wipe-manner-10.4.1} or \textit{amuse-31.1}. Each root node \(c \in \mathcal{N}_1\) from this layer represents an abstract action category, considering both the semantic role and the action target. Each node \(v \in \mathcal{N}_2\) in the second layer corresponds to a distinct verb instance within the same VerbNet class. These instances capture subtle differences in context and behavioral patterns, while sharing the same semantic roles defined by VerbNet. The third layer \(\mathcal{N}_3\) further grounds distinct verb instances into object-centric interaction templates, focusing on the specific target objects involved in the actions. Each template is aligned with a concrete skill description, and thus \(\mathcal{N}_3\) is referred to as the skill description layer. The fourth layer consisting of leaf nodes \(\sigma \in \mathcal{N}_4\) comprises fine-grained skill slices, where each slice instantiates a skill description into a specific, visually grounded example, ensuring intra-skill visual variability.

Building upon this hierarchy, automatically annotated skill segments are used to populate and expand \ourdata{}. As shown in the bottom right of Fig.~\ref{fig:data}, we first parse the skill descriptions using a VerbNet parser, providing an initial alignment with first-layer nodes of \ourdata{}. Subsequently, we perform a top-down traversal of the hierarchy, iteratively matching annotated examples with verb instances and skill descriptions at successive levels. When no suitable match is identified, a tree expansion mechanism is invoked to create a new child node, dynamically extending the hierarchy. Finally, we transform unstructured videos into a structured and continuously growing repository of skill examples, containing 106 VerbNet classes and 1659 distinct skill descriptions. 

    
    

\paragraph{Few-shot skill implementation.}

The problem of obtaining demonstrations for newly defined skills \(\pi_i\) at deployment has been transformed into efficient example retrieval from \ourdata{}. As illustrated in Fig.~\ref{fig:skill} (a), the requested skill \(\pi_i\) is first processed through a VerbNet parser to identify the entry point, and subsequently matched against the nearest nodes at each abstraction level. At the bottom layer, if multiple samples satisfy the required skill implementation, we further assess the relevance between each candidate and the target deployment scenario using the similarities between their CLIP~\cite{radford2021learning} features, which reflect coherence in viewpoint and spatial layout. Furthermore, we discard candidates whose trajectories exhibit out-of-view movement or extended periods of inactivity.

Despite sharing similar behavioral patterns, target scenes often vary in spatial layout from sample skill slices. We bridge this gap with a constrained visual prompting method, where skill-annotated segments from \ourdata{} serve as in-context examples for the target skill $\{\pi_i, p_i\}$. To capture key elements of skill execution, we extract explicit, machine-interpretable references from these examples: \textbf{fine-grained spatial trajectories} and \textbf{high-level semantic constraints}. Specifically, the 6-DoF poses of the skill demonstrations are projected onto the corresponding camera view of the initial frame, producing 2D traces visualized as green waypoints. The extracted trajectory $\tau_e$, combined with example-view observations $\mathcal{O}_e$, provides a direct trajectory reference. In parallel, contact and waypoint constraints $\{\phi_c, \phi_w\}$ are obtained from demonstrations using off-the-shelf VLMs, converting implicit procedural knowledge into transferable semantic constraints that offer high-level guidance.  


As shown in Fig.~\ref{fig:skill} (b), trajectory references $\{\tau_e, \mathcal{O}_e\}$ and semantic constraints $\{\phi_c, \phi_w\}$ are integrated with the target scene specification $\{\mathcal{O}_i, \pi_i, p_i\}$ to generate spatial trajectories. The parameter \(p_i\) of the target skill implementation is explicitly annotated in the real-world visual observation \(\mathcal{O}_i\) using bounding boxes. The target scene is discretized in a grid-based manner similar to MOKA~\cite{liu2024moka}. We utilize GPT-4o~\cite{gpt4o} to select candidate 2D points from both the grid and target objects, which are then lifted into 3D using depth information. The target trajectory consists of a predicted contact point $c$ for the gripper to first contact with, and a sequence of waypoints $\{w_j\}_{j=1}^N$.
The procedure for generating spatial trajectories in Cartesian coordinates can be summarized as:
\begin{equation}
\{c, \{w_j\}_{j=1}^N\} = \mathcal{V} (
\{\tau_e, \mathcal{O}_e\} \;, \;
\{\phi_c, \phi_w\} \;, \;
\{\mathcal{O}_i, \pi_i, p_i\}),
\end{equation}
where the input consists of trajectory references \(\{\tau_e, \mathcal{O}_e\}\), the derived semantic constraints on contact points and waypoints \(\{\phi_c, \phi_w\}\), as well as the visual observation and parameterized skill description of the target scene \(\{\mathcal{O}_i, \pi_i, p_i\}\). The output trajectory $\{c, \{w_j\}_{j=1}^N\}$ is of variable length, enabling flexible representation of long-horizon behaviors. 


The 3D waypoints are lifted into SE(3) space by attaching orientation patterns that are uniformly sampled from skill demonstrations, which serve as the source for orientation transfer and ensure one-to-one alignment with the number of waypoints. To preserve transferable orientation patterns, we introduce local frames constructed at each waypoint. Each local frame $R_{\text{local}}$ is spanned by the movement direction (defined by consecutive waypoints) together with orthogonal basis vectors obtained via cross products and orthonormalization. Concretely, each sampled source rotation $R_{\text{src}}$ is expressed in its source local frame $R_{\text{local}}^{\text{src}}$, yielding a skill-specific orientation pattern $R_{\text{skill}}$, which is then re-expressed in the target local frame $R_{\text{local}}^{\text{tgt}}$ before being mapped back into the global frame of the target scene as $R_{\text{tgt}}$:
\begin{equation}
R_{\text{skill}} = ({R_{\text{local}}^{\text{src}}})^{T} R_{\text{src}} R_{\text{local}}^{\text{src}}, \quad
R_{\text{tgt}} = R_{\text{local}}^{\text{tgt}} R_{\text{skill}} ({R_{\text{local}}^{\text{tgt}}})^{T}.
\end{equation}
The sampled rotation matrices are paired with the 3D waypoints, forming a sequence of executable 6-DoF poses.




\begin{figure*}[t]
    \centering
    \includegraphics[width=\linewidth]{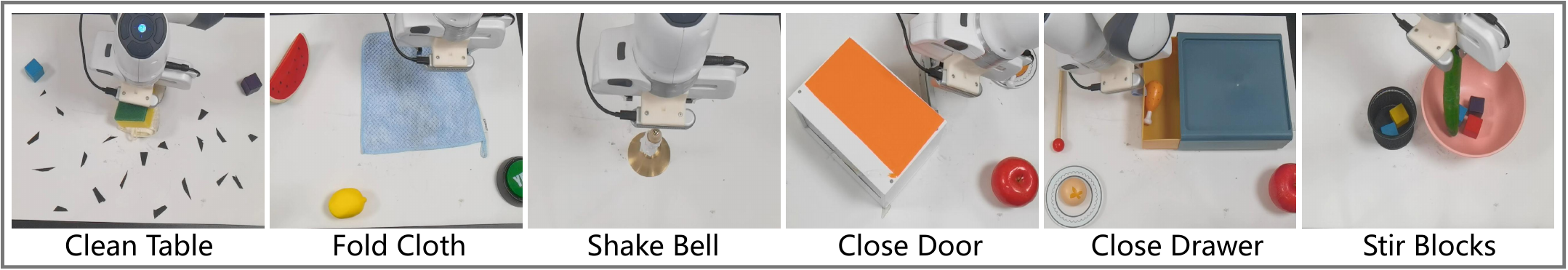}
    \caption{Example scene arrangement for real-world tasks.}
    \vspace{-1.0em}
    \label{fig:real}
\end{figure*}

\section{EXPERIMENT}
To comprehensively evaluate the performance of \our{} on manipulation tasks, we conduct both simulation and real-world experiments. The baselines contain two groups of skill-centric methods based on foundation models: (1) policy code generation methods, represented by Code-as-Policies (CaP)~\cite{cap}, and (2) visual prompting methods, represented by MOKA~\cite{liu2024moka}. We implement CaP with GPT-3.5~\cite{ouyang2022training} and GPT-4o~\cite{gpt4o} to enhance visual reasoning, and MOKA with GPT-4o following their keypoint selection strategies. To ensure fair comparison, all methods (CaP, MOKA, and \our{}) share the same basic skill set $L_{\text{base}}$, based on AnyGrasp~\cite{fang2023anygrasp} for grasping and a top-down placement policy. Tasks are categorized into two groups: those fully covered by $L_{\text{base}}$, focusing on planning and instruction understanding, and those requiring generalization beyond the basic skills. 

\begin{table}[t]
\centering
\caption{The zero-shot success rate of \our{} and baselines across 8 real-world manipulation tasks.}
    \label{tab:real}
    \begin{tabular}{l|cccc}
    \toprule
    \bf Task &  \bf CaP~\cite{gpt4o} &  \bf MOKA~\cite{liu2024moka} & \bf \our{} (ours) \\
    \midrule 
    Pick Place  & 0.80 &  0.60   &  0.90\\
    Stack Blocks & 0.20 &  0.10  &  0.50\\
    Clean Table &   0.80 &  1.00 &  1.00\\
    Fold Cloth & 0.00 & 0.20 &  0.70\\
    Shake Bell & 0.50 &  0.60  &  0.60\\
    Close Door & 0.00 &  0.10  &  0.80\\
    Close Drawer  & 0.00 &  0.50   &  0.70\\
    Stir Blocks & 0.00 & 0.00  &  0.60\\
    \bf Average & 0.00 & 0.39  &  \bf 0.73\\
    \bottomrule
    \end{tabular}
    \vspace{-1.0em}
\end{table}

\subsection{Simulation Experiments}
\label{exp:sim}
We use RLBench~\cite{rlbench} as the simulation platform, selecting 8 tasks solvable with the base skill set and 10 additional tasks requiring skill extension. Following PerAct~\cite{peract}, each task is evaluated over 25 episodes with binary success/failure scores, and the process is repeated three times to report mean and standard deviation of success rates. We first assess the performance of \our{} on tasks sampled from the pre-defined skill distribution. As shown in Table~\ref{tab:rlbench}, when task instructions are ambiguous and rely on visual grounding, \our{} demonstrates clear improvements over CaP. For instance, in the \textit{Close Jar} task, the instruction \textit{"close the jar"} necessitates interpreting the spatial context to determine the appropriate action. \our{} correctly infers that the gray lid must be placed and rotated onto a specific jar, whereas CaP often misinterprets the scene, assuming the lid is already aligned. As discussed in Sec.~\ref{sec:plan}, this advantage stems from \our{}'s skill-aware planning module, which aligns free-form instructions with visual observations.

To evaluate the adaptability of \our{} to novel tasks, we selected 10 additional tasks beyond the pre-defined skill distribution, covering six primary skill categories: \textit{Revolute}, \textit{Prismatic}, \textit{Flip}, \textit{Unplug}, \textit{Lift}, and \textit{Pour}, where the first two denote operations involving revolute or prismatic joints.\footnote{The correspondence: Revolute (Close Micro, Close Fridge, Seat Down, Close Laptop, Open Door), Prismatic (Close Drawer), Flip (Press Switch), Unplug (Unplug Charger), Lift (Lift Number), Pour (Water Plants).} As shown in Table~\ref{tab:rlbench_extra}, the policy code generation method CaP fails to generalize beyond the pre-defined skills without extra manually-annotated APIs at deployment. In contrast, \our{} incorporates skill sufficiency into its planning process and leverages diverse skill examples retrieved from \ourdata{} to implement self-augmented skills. 

As shown in Table~\ref{tab:rlbench_extra}, \our{} outperforms MOKA on a wide range of trajectory-based tasks out of the pre-defined skills. This improvement stems from the incorporation of skill-centric demonstrations retrieved from \ourdata{}, which offer both semantic constraints and referential guidance during skill execution. As a result, \our{} generates trajectories that are often more semantically meaningful and physically feasible, particularly for task categories such as \textit{Revolute} and \textit{Prismatic}. Meanwhile, we observe that MOKA frequently fails to complete compositional tasks, such as \textit{Unplug} and \textit{Lift}. Unlike MOKA, which relies on repeated interactions with VLMs to produce plans, \our{} employs a structured reasoning process grounded in a unified VLM. This enables our method to adapt to more complex scenarios that require both the decomposition of long-horizon instructions and generalization across diverse skills.

\subsection{Real-World Experiments}

As shown in Fig.~\ref{fig:real}, we further deploy \our{} in real-world settings using a Franka Emika robot arm. Two groups of tasks, covering eight diverse categories, are designed for evaluation. The first group includes tasks solvable with predefined skills, such as \textit{Pick-Place} and \textit{Stack Blocks}, aligned with the simulation tasks. The second group assesses the model's generalization to a broader range of real-world tasks, including both scenarios that resemble the simulation environment and novel, trajectory-focused tasks not included in RLBench, such as \textit{Fold Cloth} and \textit{Stir Blocks}. These more complex tasks demand reasoning about tool use, physical properties, and trajectory inference from free-form instructions. Each task is evaluated over 10 trials, with the success rate serving as the evaluation metric. To ensure robustness, object instances and layouts are reconfigured after each trial.

As shown in Table~\ref{tab:real}, owing to the robust reasoning capabilities of code-based policies, \our{} and CaP demonstrate superior performance on long-horizon pick-and-place tasks, such as compositional pick-place and block stacking. For trajectory-based tasks that require spatial reasoning over execution trajectories, such as \textit{Fold Cloth}, \our{} exhibits more stable performance than MOKA, benefiting from retrieved skill-centric examples. In more challenging tasks demanding both compositional and spatial reasoning over trajectories, such as \textit{Stir Blocks}, the advantages of \our{}'s integration of long-horizon planning and self-augmented skill composition become more pronounced. Unlike MOKA, which relies on unstructured reasoning, and CaP, which operates with a fixed skill library, \our{} decomposes this long-horizon task into subtasks, including tool localization, grasping, insertion, and a self-augmented stirring procedure. Supervised by relevant skill examples from \ourdata{}, \our{} performs multiple rounds of precise rotational stirring motions at the designated target location.

\subsection{Ablation}
\begin{figure}
    \includegraphics[width=0.47\textwidth]{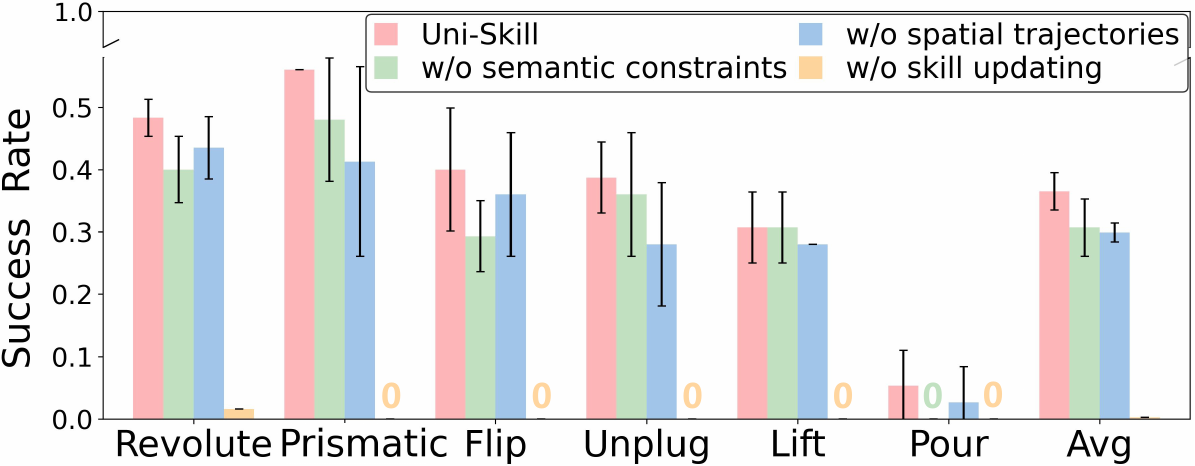}
    \caption{Ablation Results on RLBench.}
    \label{fig:ablation}
\end{figure}

\begin{table}[t]
\centering
\caption{Comparison on different raw data sources.}
    \label{tab:data_comp}
    \begin{tabular}{l|cccc}
    \toprule
    \bf Data Source &  \bf VerbNet class &  \bf Skill count & \bf Success Rate \\
    \midrule 
    DROID~\cite{droid}  & 106 &  1659  &  0.41$\pm0.01$\\
    sth2sth~\cite{goyal2017something} & 122 &  1432  &  0.29$\pm0.02$\\
    \bottomrule
    \end{tabular}
    \vspace{-1.0em}
\end{table}

We conduct ablation studies focusing on two key aspects: (1) the mechanism for updating and defining new skills in the skill-aware planning module, and (2) the referential role of examples retrieved from \ourdata{} in guiding automatic skill implementation. Similar with Sec.~\ref{exp:sim}, we conduct experiments on six categories of tasks out of the predefined skills to evaluate the performance of \our{} under different ablation settings. The skill updating mechanism is the most critical component at the top of the entire process, determining whether the system can generalize beyond predefined skills. As shown in Fig.~\ref{fig:ablation}, when the skill updating mechanism is disabled, the system fails to perform reasonably on most of tasks. For components of the automatic skill evolution module, we further investigate the respective roles of semantic constraints and spatial trajectories extracted from relevant skill examples in few-shot skill implementation. As shown in Fig.~\ref{fig:ablation}, the relative importance of each component varies across task categories. Tasks involving contact-sensitive interactions and requiring holistic behavioral constraints (e.g., \textit{Revolute}, \textit{Flip}) demonstrate a more significant performance degradation when semantic constraints are omitted from the in-context input. In contrast, tasks that require more precise spatial reasoning (e.g., \textit{Unplug}, \textit{Lift}) are more affected by the removal of spatial trajectory references. These two types of referential information serve distinct roles within the in-context inputs, effectively supporting few-shot skill implementation without manual intervention.

\subsection{Discussion on adaptation and robustness}
\textbf{Adaptation to ego-centric videos.}
We use DROID~\cite{droid} as the source of unstructured videos, which is robot-centric and the most relevant. However, our automatic annotation pipeline is not restricted to robotic demonstrations. With lightweight hand detection modules, skill-centric segments can be effectively extracted from ego-centric videos like sth2sth~\cite{goyal2017something}. We report summarized VerbNet classes, skill count and average success rate on 10 RLBench tasks with different raw data sources. Compared with robotic data, human-centric videos cover a broader and more diverse range of skill categories. However, human-centric videos primarily suffer from the absence of action annotations and contain blurred frames, which degrades the quality of generated skill repositories. Consequently, human-centric Internet videos are promising sources for scaling of skill demonstrations, though requiring further quality control and data filtering.

\begin{figure}
\centering
    \includegraphics[width=0.38\textwidth]{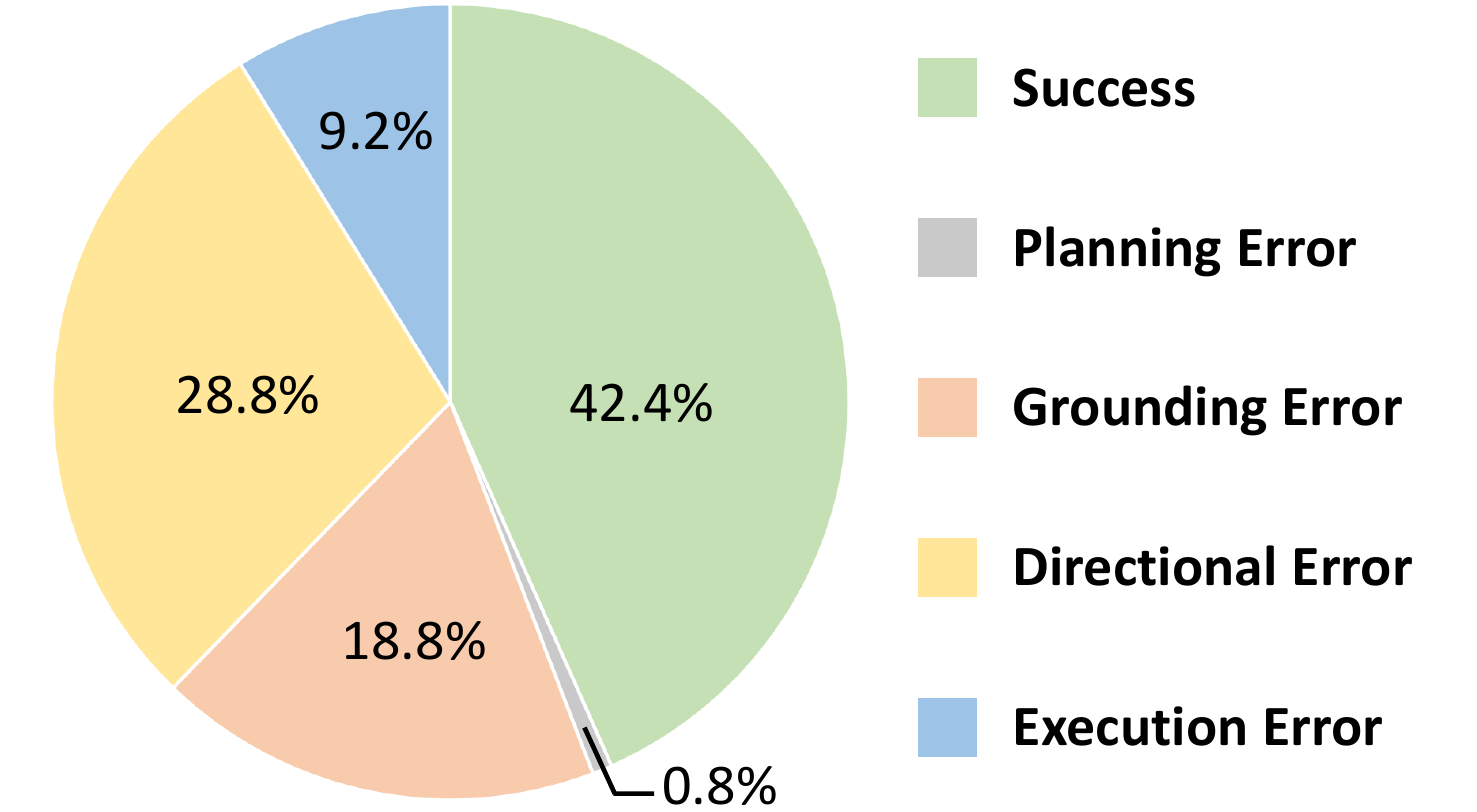}
    \caption{Failure modes of \our{} out of the basic skill set.}
    \vspace{-1em}
    \label{fig:failure}
\end{figure}

\textbf{Failure mode and recovery.} 
As shown in Fig.~\ref{fig:failure}, we analyzed modes of failure cases of experiments on tasks out of the pre-defined skills. The failure modes fall into four types: planning errors, grounding errors, directional errors, and execution errors. Planning errors from task decomposition (e.g., missing intermediate step) are largely mitigated through training with multi-modal aligned code data. Execution errors originate from the inverse kinematics module of RLBench or disabilities of basic APIs (e.g., IK divergence near joint limits). Grounding errors stem from the perception module or viewpoint occlusions (e.g., misidentifying the target handle), and are especially common in tasks requiring subtle visual discrimination. Advances in open-vocabulary detection methods are expected to alleviate these issues. Directional errors occur when visual prompting provides incorrect movement guidance, often due to mismatches between sample and target trajectories or VLM hallucinations. (e.g., moving forward instead of downward). To address these mismatches, we introduce a self-correcting mechanism that enforces a closed-loop process. Failure cases are incorporated as negative in-context examples, enabling the model to diagnose the error cause and re-plan the trajectory. This strategy proves effective across diverse tasks; for instance, it further improves the success rates on \textit{Close Drawer} and \textit{Seat Down} by 12\%.

\textbf{Annotation quality.} 
To have an explicit evaluation on data quality annotated by an automatic VLM pipeline, we performed quality assessment for invalid slices on \ourdata{}. Out of 135 skill slices sampled from \ourdata{}, only 2 showed semantic mismatches, 7 had occlusions or were too short, and zero execution failures. With filtering on trajectory lengths and boundaries, we largely eliminated the second type of issue, reducing low-quality data to under 2\%. Proper filtering and quality control can largely mitigate the noise in unstructured robotic demonstrations. 

\section{Conclusion and Future Work}
We propose \our{}, a skill-centric framework that enables zero-shot generalization out of pre-defined skills. \our{} builds on the skill-aware planning mechanism to detect skill gaps and augment new skills if required. Through our hierarchical skill repository, \ourdata{}, we effectively bridge extensive automatically annotated demonstrations with structured skill taxonomy. After efficient retrieval from \ourdata{}, skill-centric examples further provide guidance for \our{} to perform diverse real-world tasks and generalize to unseen skills. In this work, we use semantic hierarchies to organize skills and retrieve skill-related examples. Exploring an alternative skill retrieval mechanism that jointly accounts for mechanical properties and semantic categories remains a meaningful direction for further exploration, which could further eliminate overlaps between skills and enable more effective sample retrieval.


\bibliographystyle{IEEEtran}
\bibliography{IEEEexample}

\end{document}